\documentclass[twoside,11pt]{article}

\usepackage{blindtext}

\usepackage{jmlr2e}

\usepackage{import}

\usepackage{albi_pack}

\usepackage{lastpage}
\jmlrheading{23}{2022}{1-\pageref{LastPage}}{1/21; Revised 5/22}{9/22}{21-0000}{Alberto Carlevaro and Teodoro Alamo and Fabrizio Dabbene and Maurizio Mongelli}

\ShortHeadings{Exact SDR for Exponential Distributions and MC-SVM Approximation}{Carlevaro, Alamo, Dabbene and Mongelli}
\firstpageno{1}

\begin{document}

\title{Exact characterization of $\varepsilon$-Safe Decision Regions for exponential family distributions and Multi Cost SVM approximation}

\author{\name Alberto Carlevaro \email alberto.carlevaro@ieiit.cnr.it \\
       \addr Istituto di Elettronica e di Ingegneria
dell’Informazione e delle Telecomunicazion\\
       Consiglio Nazionale delle Ricerche\\
       00185 Rome, Italy
       \AND
       \name Teodoro Alamo \email talamo@us.es \\
       \addr Escuela Superior de
             Ingenieros\\
       Universidad de Sevilla\\
       41020 Seville, Spain
       \AND
       \name Fabrizio Dabbene \email fabrizio.dabbene@ieiit.cnr.it \\
       \addr Istituto di Elettronica e di Ingegneria
dell’Informazione e delle Telecomunicazion\\
       Consiglio Nazionale delle Ricerche\\
       00185 Rome, Italy
       \AND
       \name Maurizio Mongelli \email maurizio.mongelli@ieiit.cnr.it \\
       \addr Istituto di Elettronica e di Ingegneria
dell’Informazione e delle Telecomunicazion\\
       Consiglio Nazionale delle Ricerche\\
       00185 Rome, Italy}

\editor{Pradeep Ravikumar, Carnegie Mellon University.}

\maketitle

\begin{abstract}
Probabilistic guarantees on the prediction of data-driven classifiers are necessary to define models that can be considered reliable. This is a key requirement for modern machine learning in which the goodness of a system is measured in terms of trustworthiness, clearly dividing what is \textit{safe} from what is \textit{unsafe}. The spirit of this paper is exactly in this direction. First, we introduce a formal definition of \textit{$\varepsilon$-Safe Decision Region}, a subset of the input space in which the prediction of a target (safe) class is probabilistically guaranteed. Second, we prove that, when data come from exponential family distributions, the form of such a region is analytically determined and controllable by design parameters, i.e. the probability of sampling the target class and the confidence on the prediction. However, the request of having exponential data is not always possible. Inspired by this limitation, we developed Multi Cost SVM, an SVM based algorithm that approximates the safe region and is also able to handle unbalanced data. The research is complemented by experiments and code available for reproducibility.
\end{abstract}


\section{Introduction and Problem Formulation}
\label{sec:intro}
Modern machine learning (ML) algorithms face significant uncertainty due to the inherent variability of data, the complexity of model architectures, and challenges in controlling predictions. Developing models that provide consistent and trustworthy results, regardless of data or algorithm variations, is essential \cite{kaur2022trustworthy,10.1145/3555803,liu2022trustworthy} and indeed, to prevent undesirable behavior, the artificial intelligence (AI) community has been working on the idea of mitigating this uncertainty by controlling the training sets and parameters of ML models \emph{a priori} \cite{damiani2020certified,maleki2020machine,menzies2005verification,myllyaho2021systematic,tao2019testing}. In this paper, we take a different \emph{a posteriori} approach to verify ML models by providing probabilistic certifications of the outcomes to ensure the reliability and robustness of these algorithms. Our research contributes to this by developing a (binary) support vector machine (SVM, \cite{SVM}) based classifier, the Multi Cost SVM algorithm, that offers probabilistic guarantees on its output and is robust to changes in the a priori probability of the events it predicts. The idea comes from the observation that for the special class of data coming from exponential family distributions (e.g., Gaussian, gamma, Chi-squared, etc.) it is possible to draw a ``decision region'' of the input parameters (a true classifier) in which the probability of observing a desired event is controlled by the confidence requested on the prediction regardless of the a priori probability of the training data. This idea bridges robust classification and guaranteed predictions, offering a reliable tool for data classification. Therefore, this approach contributes to the development of knowledge in many areas of machine learning classification. Quantile regression, for example, is the statistical reference to which this research belongs \cite{koenker2005quantile}. In fact, the idea of probabilistic certification is the motivation for other theories derived from quantile regression, such as probabilistic scaling (PS) \cite{carlevaro2023probabilistic, carlevaroCOPA} and conformal prediction (CP) \cite{angelopoulos2023conformal}. From PS this research retrieves the idea of having a scalar parameter (sometimes referred to as ``radius'', although this term is somewhat inaccurate since it can also take negative values) to control the output of the classifier that has a confidence in the prediction in the spirit of CP. In this context, previous work, such as \cite{PIETRASZEK2005169}, attempted to control false positives using alert-management systems without probabilistic guarantees. Others, like \cite{singh2019eth}, focus on robustness against uncertainty by abstracting possible outputs. Conformity and error control are not however the unique contributions to the reliability of the model. For example, the robustness with respect to the a priori probability of the training data also puts this theory in line with unbalanced classification \cite{9408661}. In fact, always inspired by the observation that exponential family distribution define classification regions where the effect of the prior probability is concentrated in a bias term, MC SVM is designed such that it profiles a classification function that generalizes to the a priori probability of the data source. The idea at the basis of the algorithm is inbetween ensemble learning \cite{dong2020survey} and federated learning \cite{zhang2021survey}. Specifically, the classifier is simultaneously trained with multiple weighted SVMs to optimize the same vector of learnable parameters, making the algorithm more robust and capable of dealing with data coming from different balanced source probabilities. Thanks to all these factors, this theory is prone to be used in many and various applications. Examples can be numerous too, from healthcare (e.g. detection of rare diseases \cite{schaefer2020use}) to finance (e.g. the well-known problem of unbalance classification in credit card fraud detection \cite{kalid2020multiple}) and autonomous driving (e.g. assessing reliable classification to avoid collisions \cite{muhammad2020deep}) and many others can be cited. But applications are not the main purpose of this research. As we have tried to emphasize in this introduction, the paper stands out by addressing the need for a clear definition of ``safety'' in classification and providing a robust theory adaptable to varying data conditions. We can say, indeed, our approach aligns with the broader goal of ``SafeML'', aiming to create methodologies for more reliable data-driven algorithms. 

The remainder of the paper is organized introducing the reader to the theory and its contribution in the machine learning field, then we introduce the concept of ``safety region'' for exponential family distributions, some examples follow and the paper concludes with the introduction of the MC SVM algorithm, its validation and discussion. The whole research is complemented by experiments and code available for reproducibility.
\subsection{Contribution}
We consider situations in which an external probabilistic parameter $p_S\in(0,1)$, that we name \textit{safety probability}, is known to influence the outcome of our experiments. In words, this parameter can be interpreted as the probability that a given instance of our observed point $\x\in\X$ in the parameter space belongs to a given set $S\subset\X$. The parameter $p_S$ is somewhat related to the notion of unbalanced datasets. Indeed, it is very frequent to encounter practical situations in which one observes a possibly large unbalance in binary datasets. For instance, in a production line (imagine, for instance, a line producing electronic circuits) one usually encounters a large number of good situations (i.e. ``safe'' production) and -- hopefully -- a few cases of defective productions (i.e. bad chips).  
Moreover, the ratio between safe and total number of products (which indeed is governed by $p_S$, at least in expectation) may vary among different production lines, due to different external factors (e.g.\ quality of the prime materials, workforce).
In such cases, it is natural to assume that the safe and unsafe products obey the same distribution, and the changes/unbalances are modeled by the safety probability level.
Formally, we may assume that the density of $\x\in\X$ is governed by the (Bernoulli) parameter $p_S$ as follows:
\begin{equation}
    \label{eq:f_x1}
f(\x) = f(\x|S)p_S + f(\x|U)(1-p_S),
\end{equation}
where $U$ is the complementary event of $S$ and $f(\x|S)$, $f(\x|U)$ are the densities of the safe and unsafe points.

Motivated by the above considerations about the importance of obtaining classifiers able to provide provable safety guarantees, 
in this work we introduce a novel definition of \emph{$\varepsilon$-Safe Decision Region} ($\varepsilon$-SDR) as follows:
\begin{equation}
    \label{eq:PCR}
\Phi_\varepsilon = \left\{\x : p(S|\x) \ge 1-\varepsilon\right\},
\end{equation}
where $p(S{\mid}\x) \in [0,1]$ denotes the conditional probability (function) that calculates the probability that a certain event $S$ occurs given the observation of an instance $\x$.
It outlines the set of samples $\x$ such that the probability of observing the event~$S$ given $\x$ is greater or equal than $1-\varepsilon$, where $1-\varepsilon$ denotes the confidence (and thus $\varepsilon$ the error, or \textit{risk level}). In words, a point in $\varepsilon$-SDR has a high probability (at least $1-\varepsilon$) of being safe.
Clearly, for a given problem, the set $\varepsilon$-SDR will depend on the confidence level $1-\varepsilon$:  one would expect that the smaller is the risk level $\varepsilon$
the smaller will be the set $\Phi_\varepsilon$.
More importantly, the size of the $\varepsilon$-SDR will also be critically influenced by the safety probability $p_S$. Indeed, also in this case one should expect that when $p_S$ decreases the size of the ensuing $\varepsilon$-SDR would also decrease. 

The contribution of this paper is twofold. First, for a very general class of distributions, namely the so-called exponential family of densities \cite{abramovich2013statistical}, we prove that the $\varepsilon$-SDR depends in a very precise way on the safety probability parameter $p_S$ and on the risk value $\varepsilon$. Indeed, the $\varepsilon$-SDR presents a boundary $\Gamma(\x) = \rho$, uniquely determined by the training points, while the probability $p_S$ and the risk $\varepsilon$ only change the size of the set through the radius $\rho(p_S,\varepsilon)$. That is, we have
\begin{equation*}
    \Phi_\varepsilon = \left\{\x : \Gamma(\x) \le \rho(p_S,\varepsilon) \right\},
\end{equation*}
where the function $\Gamma(\x)$ is independent of $p_S$ and $\varepsilon$ and determines the ``shape'' of the $\varepsilon$-SDR, while $\rho(p_S,\varepsilon)$ simply selects the appropriate sublevel set of $\Gamma(\x)$ .

This key fact, which \textit{per se} represents an important theoretical result, motivates our second contribution. Indeed, the consideration that, for data with generic exponential distributions, the ``shape'' of the safety region is independent of the underlying probability of safety (and, thus, in a sense, on the way the data is balanced), means that a good classifier shall also not depend on this aspect. 
Hence, in the second part of the paper we develop a novel algorithm for designing SVM binary classifiers that is also able to handle discrepancy in data balance, the already cited Multi Cost SVM (see Section \ref{sec:MCSVM}). This algorithm is a first rigorous approach to design a classifier that is independent of the a priori probability of the training data. The idea is to solve an ensemble of SVMs with different weighting parameters to obtain a classification boundary independent of the data probability, i.e. general for any level of probability $p_S$. Then, to obtain a safe guarantee on the prediction, the offset of the model is appropriately tuned (see Section \ref{subsec: on the choice of b}).
\protect{\remark[On the experiments and the spirit of the paper]{The focus of this research is mainly theoretical. Consequently, all the experiments and examples that will be shown in the following did not require sophisticated hardware or long training times. For reproducibility, the Matlab code of the experiments can be downloaded at \url{https://github.com/AlbiCarle/Multi-Cost-SVM} and can be easily run on ordinary laptops}}.
\section{Safety Regions for Exponential Families}
\label{sec:safety_regions_exp_families}
We consider two complementary events $S$ and $U$, with probabilities $p_S$ and $p_U=1-p_S$. To give them a representative meaning, we can consider them respectively as ``Safe'' and ``Unsafe'' configurations, respectively.
We consider then a sample space $\mathcal{X}$ with probability density function $f(\x)$. We assume that the events $S$ and $U$ have a probabilistic effect on $\x$, that is, we consider the density functions $f(\x|S)$ and $f(\x|U)$ that serve to characterize $f(\x)$ in terms of $S$ and $U$ to be expressed as
\begin{equation}
\label{eq:f_x}
f(\x) = f(\x|S)p_S + f(\x|U)p_U.    
\end{equation}
The main assumption of this work is that we consider $f(\x|S)$ and $f(\x|U)$ belonging to an exponential family \cite{abramovich2013statistical}, i.e., for all $\x\in \mathcal{X}$, we can write
\begin{equation}
\label{eq:exponential}
\begin{split}
f(\x|S) = c_S^{-1}\exp\left(-g_S(\x)\right), 
\quad
f(\x|U) = c_U^{-1}\exp\left(-g_U(\x)\right),
\end{split}
\end{equation}
\noindent where
 $g_S(\cdot)$ and $g_U(\cdot)$ are measurable functions of $\x$.
 The normalising constants $c_S$ and $c_U$ are given by
\begin{eqnarray}
c_S = \int\limits_\mathcal{X} \exp\left(-g_S(\x)\right) d\x,
\quad
c_U = \int\limits_\mathcal{X} \exp\left(-g_U(\x)\right) d\x.
\end{eqnarray}
Examples of distributions that belong to this family are Gaussian, Chi-squared, gamma, Poisson, Bernoulli, while Cauchy, hypergeometric and Student's t are examples of distributions that do not belong to this family.

Denote $p(S|\x)$ the probability of event $S$ for a given point $\x$. We now focus on the notion of \emph{safety region}  \cite{SetoAnEngineering1999,9594676}, i.e. the region $\Phi_\varepsilon\subseteq\mathcal{X}$ for which the probability of event $S$ given $\x$ is larger than $1-\varepsilon$, where $\varepsilon\in(0,1)$ is a probabilistic parameter representing the error. 
%
\begin{definition}[$\varepsilon$-Safe Decision Region]
\label{def: PCR}
Given the density function $f(\x)$ as in \eqref{eq:f_x} and the risk level $\varepsilon \in (0,1)$, we define the $\varepsilon$-Safe Decision Region ($\varepsilon$-SDR) related to $S$ as the region in which the probability of observing the ``safe'' event S has a level of confidence of at least $1-\varepsilon$, i.e.
\begin{equation*}
 \Phi_\varepsilon \doteq \set{\x}{p(S|\x) \ge 1-\varepsilon}
= \set{\x}{p(U|\x)<\varepsilon}.   
\end{equation*}
\end{definition}
The above definition is very theoretical and can be applied to any probability distribution. However, if we consider the exponential hypothesis about probabilities, we find that $\Phi_\varepsilon$ has a more ``operational'' form in which it becomes clear that these regions have a uniquely determined boundary shape that is controlled by a parameter that functions as a radius.\footnote{In fact, one should speak of a ``meta-radius'' because, as will become clear in the following, it can also take on negative values.} The next proposition formalizes these concepts.
\begin{proposition}[Scaling form of $\varepsilon$-SDR]\label{PropositionBella}
Assume that $f(\x|S)$ and $f(\x|U)$ are of the exponential form  \eqref{eq:exponential}, and that the points $\x$ obey density \eqref{eq:f_x} for given $p_S\in (0,1)$. Then, for given risk level $\varepsilon\in(0,1)$, the $\varepsilon$-SDR can be written as 
    \begin{equation*}
    \label{equ:Phi:A:exponencial}
    \Phi_\varepsilon = \set{\x}{ \Gamma(\x) \le \rho(p_S,\varepsilon)} 
\end{equation*}
where
\begin{eqnarray}
\Gamma(\x)&\doteq&g_S(\x)+\ln c_S-g_U(\x)-\ln{c_U}\label{eq:Gammax} \nonumber\\
&\doteq&\tilde{g}_S(\x) -\tilde{g}_U(\x),
\label{eq:tilde_g}\\
\rho(p_S,\varepsilon) &\doteq&\ln\frac{p_S} {1-p_S} + \ln \frac{\varepsilon} {1-\varepsilon} \nonumber\\
&\doteq&\rho_{p_S} + \rho_\varepsilon.
\label{eq:radius}
\end{eqnarray}
\end{proposition}

The proof of Proposition \ref{PropositionBella} is given in the Appendix, Section \ref{Proof: PropostionBella}.

As already discussed in the introductory section, Proposition \ref{PropositionBella} means that the $\varepsilon$-SDR for exponential distributions is controllable by an offset depending on probabilistic parameters $p_S$ and $\varepsilon$. That is, once its shape has been uniquely determined by the observations $\x$ through $\Gamma(\x)$, it is sufficient to vary the offset $\rho(p_S,\varepsilon)$ for getting the desired amount of confidence $1-\varepsilon$ of being in $\Phi_\varepsilon$, given probability $p_S$. 

Moreover, we see that the effects of $p_S$ and $\varepsilon$ are mutually independent, and that they both act as expected: indeed, the radius increases when $p_S$ and $\varepsilon$ increase (and, somehow surprisingly, through the same  function $\ln\frac{x}{1-x}$).

To make things clearer, the following example 
focuses on a specific exponential family, the Gaussian distribution, 
showing how we can give an exact description of the $\varepsilon$-SDR in such case.

\begin{example}[$\varepsilon$-SDR form for Gaussian distribution]
\label{example:PCRgaussian}
\begin{figure*}[h!]
    \centering
    \includegraphics[width=4.5in]{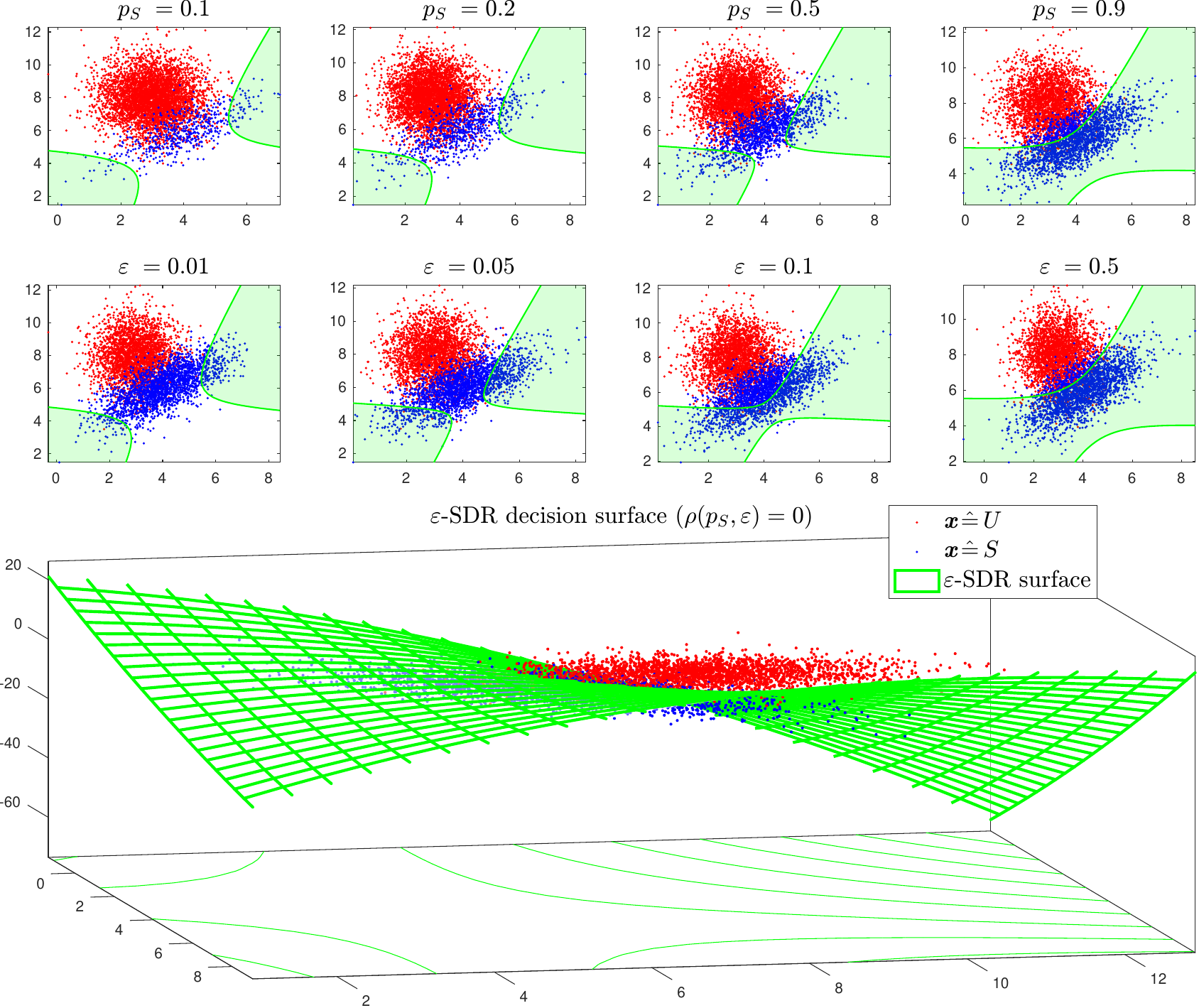}
    \caption{$\varepsilon$-SDR for Gaussian distribution. In the first row, the $\varepsilon$-SDRs at fixed $\varepsilon = 0.5 \; (\rho_\varepsilon = 0)$ are plotted as the probability of sampling safe points varies. In the second one instead the probability of sampling is fixed at $p_S = 0.5 \; (\rho_{p_S} = 0)$ and the confidence varies. Below them the function $\Gamma(\x)$ is plotted together with its level sets (that correspond to different decision boundaries, i.e. different $\varepsilon$-SDRs). }
    \label{Gaussian PCR}
\end{figure*}
Figure \ref{Gaussian PCR} shows the behavior of the $\varepsilon$-SDR as $p_S$ and $\varepsilon$ vary, respectively, for normal samples coming from these means and covariance matrices: 
$$\boldsymbol{\mu}_S = [4,6]^\top, \Sigma_S = \begin{bmatrix}
1.3 & 0.9\\
0.9 & 1.3 
\end{bmatrix}; \quad
\boldsymbol{\mu}_U = [3,8]^\top, \Sigma_U = \begin{bmatrix}
0.6 & 0\\
0 & 1.4
\end{bmatrix}.
$$
As reported in the Appendix \ref{appendix:PCRgaussian}, it is possible to completely characterize the $\varepsilon$-SDR for the Gaussian distribution from the matrix of the difference between the inverses of covariance matrices. For this example, $\Sigma_S^{-1} - \Sigma_U^{-1} = \begin{bmatrix}
-0.19 & -1.02\\
-1.02 & 0.76 
\end{bmatrix}$ has eigenvalues of different sign, which implies that in this case the decision border of the $\varepsilon$-SDR is an hyperbola. The choice of taking $p_S = 0.5$ and $\varepsilon = 0.5$ leads respectively to  $\rho_{p_S} = 0$ and $\rho_{\varepsilon} = 0$.
\end{example}
This example (and in particular Figure \ref{Gaussian PCR} is quite explanatory) gives us the opportunity to focus on the fact that the $\varepsilon$-SDR for the exponential distribution is controllable by the radius $\rho(p_S,\varepsilon)$ and that its shape does not depend on the values of the probability $p_S$ and confidence $\varepsilon$ but only on the data $\x$. Thus, fixed the data, the shape is determined by $\Gamma(\x)$.
More can be said about the role of $p_S$ and $\varepsilon$. Thanks to the property that $\rho(p_S,\varepsilon) = \rho_{p_S} + \rho_{\varepsilon}$, the contributions of the parameters are independent of each other. Thus, for a fixed probability $p_S$, the $\varepsilon$-SDR can be controlled solely by confidence or vice versa. Again from Figure \ref{Gaussian PCR}, it is easy to confirm that smaller values of the parameters $p_S,\varepsilon$ mean smaller $\varepsilon$-SDRs: if an observation $\x$ corresponding to a safe point is sampled with low probability $p_S$ or that $\Phi_\varepsilon$ is defined for an high confidence $1-\varepsilon$ (thus, small error $\varepsilon$), the price to be paid is that the $\varepsilon$-SDR could be small. In a sense, we are trying to minimize false positives, that is, those observations that belong to event $U$ but are actually classified in $S$. Of course, smaller the region, lower the probability of making errors. This comment allows us to introduce readers to the fact that the $\varepsilon$-SDR can be learned when the distribution is not known. In other words, the problem of estimating $\Gamma(\x)$, which determines how the $\varepsilon$-SDR changes with specific values for $p_S$ and $\varepsilon$, can be addressed by exploiting information coming from the data. In particular, as we will see later in the discussion, once the shape of the $\varepsilon$-SDR is learned, it is sufficient to vary the radius to scale the region to the desired confidence or probability $p_S$.
\section{SVM based approximations of the safety region}
\label{Sec: MC-SVM}

Suppose that we have a collection of labelled data points 
\[\mathcal{Z}\doteq
\left\{(\x_i,y_i)\right\}_{i=1}^n,
\]
where 
$$ y_i = \bsis{l} +1  \text{ if } \x_i \text{ has label } S, \\ -1  \text{ if } \x_i \text{ has label } U.  \esis $$ 
Sometimes in the text, we will use the notation ``$i\in[n]$'' to indicate that the index $i$ varies along all the integers from $1$ to $n$, i.e. $i=1,\dots, n$. We assume that the data in $\mathcal{Z}$ has been generated from the distribution in \eqref{eq:f_x}.
Also, we define the two subsets $\mathcal{Z}^+\subseteq \mathcal{Z}$ and $\mathcal{Z}^-\subseteq \mathcal{Z}$ respectively as the subset of data with positive labels (safe data) and the subset of data with negative labels.

\begin{remark}[On $p_S$ and data unbalance]
We note that there exists a probabilistic correlation between the safety probability $p_S$ underlying the data-generation mechanism (according to the data distribution \eqref{eq:f_x}) and the observed data unbalance between the two classes. Indeed, it is easy to see that the expected cardinality of the set $\mathcal{Z}^+$ is just $n\cdot p_S$. In other words, the lower the probability of generating a safe sample is, the more the set will be unbalanced towards the negative class, and viceversa.
    
\end{remark}

Our goal is to find a good approximation of the safety region $\Phi_\varepsilon$ when we do not know the distribution of the data. To this end, we introduce a parameterization of $\tilde g_S(\x)$ and $\tilde g_U(\x)$ in \eqref{eq:tilde_g} as follows:
\[
\tilde g_S(\x)  \approx  \w_S^\top \varphi(\x), \quad
\tilde g_U(\x)  \approx  \w_U^\top \varphi(\x), 
\]
where $\varphi$ is some feature map (possibly nonlinear).
We notice from \eqref{eq:tilde_g} that the expression for the safety region depends on the difference of $\tilde g_S(\x)$ and $\tilde g_U(\x)$, that is, on $(\w_S-\w_U)^\top \varphi(\x)$.
Thus, in order to obtain/estimate the safety region it suffices to consider the difference $\w=\w_S-\w_U$: 
\begin{eqnarray*}
    \Phi_\varepsilon&\approx& \tilde\Phi_\varepsilon \doteq \set{\x}{\w^\top \varphi(\x)   \leq  \rho(p_S,\varepsilon) }.
\end{eqnarray*}

Then, defining 
$ b = \rho(p_S,\varepsilon)$,
we have
\begin{equation}
\tilde{\Phi}_\varepsilon = \set{\x}{\w^\top \varphi(\x) - b \leq 0}.
\label{equ:Phi:w}
\end{equation}
That is, the approximate $\varepsilon$-SDR $\tilde{\Phi}_\varepsilon$ can be represented by the decision region of an SVM (but other representations can be used, e.g. neural networks). In essence, it is a set of points classified as belonging to the ``Safe'' ($y = +1$) class.
%
In the next section, we show how to construct $\w$. The idea is that the uncertainty brought by different values of probability of observing the safe class $S$ can be handled through ensambling \emph{multiple} SVM predictors. Then, the design of $b$ allows to recover the level of confidence of the prediction, obtaining at the very end, a classifier capable of approximating sufficiently well the $\varepsilon$-SDR.
\subsection{Multi Cost SVM}
\label{sec:MCSVM}
We now present a first (naive) approach for the computation of $ \mathbf{w}$ and  $b$, based on classical SVM. We would like $\w^\top \varphi(\x_i) - b$ to be negative if $\x_i$ is labelled as $S$ (i.e. $y_i=+1$), and positive otherwise. That is, we would like the quantity $y_i(\w^\top \varphi(\x_i) - b)$ to be negative with high probability in every situation.
Based on this, we now recall the classic SVM optimization problem \cite{SVM}, leading to a classifier distinguishing between classes $S$ and $U$:
\begin{eqnarray*}  &\min\limits_{\w,b, \xi_1,\ldots, \xi_n} & \frac{1}{2\eta}\w^\top \w + \Sum{i=1}{n} \xi_i  \\
& \text{s.t.}& y_i(\w^\top \varphi(\x_i) - b) \leq \xi_i-1,\\ 
& &\xi_i \geq 0, \; i \in 
[n].
\end{eqnarray*}
The hyper-parameter $\eta>0$ serves to make a trade-off between regularization and misclassification error. 
Once the value of $\w$ and $b$ have been obtained, we could provide the following classifier:
$$ \hat{y}(x) = \bsis{ll} +1 & \text{ if } \w^\top \varphi(\x) - b < 0 \\ -1 & \text{ otherwise}. \esis  $$
Let us now denote as {\it false positive} the situation in which $\w^\top \varphi(\x_i) -b$ is negative when the label of $\x_i$ is $U$ and {\it false negative} when $\w^\top \varphi(\x_i) -b$ is non-negative and $\x_i$ is labelled $S$. 

The previous formulation penalizes in the same way both misclassification errors. In order to cope with this, we introduce a weighting parameter $\tau\in (0,1)$, and  we formulate the \textit{weighted SVM} (w-SVM) problem
\begin{eqnarray}
\label{eq:SVM_weighted}
&\min\limits_{\w,b , \xi_1,\ldots, \xi_n } & \frac{1}{2\eta}\w^\top \w + \frac{1}{2}\Sum{i=1 }{n} \left((1-2\tau)y_i +1\right)\xi_i \nonumber \\
& \text{s.t.} & y_i(\w^\top \varphi(\x_i) - b) \leq \xi_i-1, \; i\in[n],\\ 
& &\xi_i \geq 0, \; i\in[n]. \nonumber 
\end{eqnarray}
We note that this approach is somehow standard in SVM for unbalanced sets, and follows the same philosophy of~\cite{SVM_Veropoulos}, which suggests using different error costs for the positive and negative classes.
In~\cite{SVM_Veropoulos} this choice was motivated by the observation that the separating hyperplane of an SVM model developed with an imbalanced dataset can be skewed towards the minority class, and this skewness can degrade the performance of that model with respect to the minority class.
But there is no defined indication of what values should be preselected as the penalty parameters and the choice is totally empirical. The w-SVM  method is referred to as Different Error Costs SVM (DEC-SVM) in \cite{Class_imbalance_survey} and Cost-Sensitive SVM in \cite{CS-SVM}.

Indeed, while $\tau=0,5$ would weight equally the two classes, it is clear that values of $\tau$ close to zero will penalize quite a lot the false negative errors and very little the positive errors. On the other hand, values of $\tau$ close to one yield just the opposite behavior. The value of $\tau$ is then related to the false positive rate. As a matter of fact, it is well known in the quantile regression literature \cite{koenker2005quantile}, that if one discards the regularization term (i.e. $\eta\to \infty$) then the false negative ratio tends to $\tau$ when the number of samples tends to infinity (under not very restrictive assumptions on the data). However, when $\eta$ is not a large value, it is not that simple to relate $\tau$ with the false negative ratio. Only a qualitative relationship exists. From the previous discussion we infer that the optimal parameter $\w$ does not depend on the particular specifications for $\varepsilon$, or the value of $p_S$. Thus, in our opinion a reasonable scheme would be to obtain the value of $\w$ that better fits different values of $\tau$ (see example \ref{ex:multiple_tau}). Specifically, an optimization problem is formulated considering different values $\tau_k$ with the corresponding optimal values $b_k$ but all of them sharing the same value for $\w$. 
That is, given $\eta>0$ we introduce the set of positive values $\mathcal{T} \doteq \{\tau_1,\tau_2,\ldots, \tau_m\}$.
Then, for each $i=1,\dots, n$ and $k = 1,\dots,m$, we let
\begin{equation*}
\xi_{i,k} \doteq \max\bigl(0, y_i(\w^T\varphi(\x_i)-b_k)\bigr),
\label{def: xi}
\end{equation*}
and thus we introduce the decision variables $\w$, $\mathbf{b}$, and $\xiB_1,\ldots, \xiB_n$, with
\begin{eqnarray}
\mathbf{b} &\doteq&
\left[
\begin{array}{cccc}
b_1 &  b_2 & \cdots &  b_m 
\end{array}
\right]^\top,\\
\xiB_i &\doteq&
\left[
\begin{array}{cccc}
\xi_{i,1} &  \xi_{i,2} & \cdots &  \xi_{i,m} 
\end{array}
\right]^\top, \; i\in [n].    
\end{eqnarray}
The resulting optimization problem is
\begin{equation}  
\begin{split}
&\min\limits_{\w,\mathbf{b}, \xiB_1,\ldots, \xiB_n}  \frac{1}{2\eta}\w^\top \w \ +\\
&\hspace{1.8cm} \frac{1}{2}\Sum{k=1}{m} \Sum{i=1 }{n} \left((1-2\tau_k)y_i +1\right) \xi_{i,k} \\
& \quad \quad \text{s.t.} \quad \quad y_i(\w^\top \varphi(\x_i) -b_k) \leq \xi_{i,k} -1 , \\ 
& \quad \quad \quad \quad \quad \ \xi_{i,k} \geq 0, \; i\in[n], k\in[m].
\end{split}
\tag{MC-SVM}\label{eq: opt_prbl_primal}
\end{equation}
In words, we consider an optimization problem composed by $m$ SVMs with the same data $\{(\x_i,y_i)\}_{i=1}^n$, with a unique hyperplane $\w$ but different offsets $b_k$ corresponding to different weights $\tau_k$, $k\in[m]$. From now on, we will refer to the model \eqref{eq: opt_prbl_primal} as \textit{Multi Cost SVM} (MC-SVM), to convey the idea of reliability and robustness in the model's predictions across different scenarios. 

In order to derive the Lagrangian dual of \eqref{eq: opt_prbl_primal} we define:
\begin{eqnarray} 
R &\doteq& \bmat{cccc} \varphi(\x_1) & \varphi(\x_2) & \ldots & \varphi(\x_n)\emat^\top,\label{eq:R} \\ 
D&\doteq&\text{diag}\{ y_1, y_2,\ldots, y_n\},\label{eq:D} \\
\boldsymbol{1} &\doteq&\left[
\begin{array}{cccc}
1 &  1 & \cdots &  1 
\end{array}
\right]_{1\times n}^\top, \label{eq:ones}\\
K&\doteq&RR^\top, \label{eq:K}
\end{eqnarray}
where, introducing the \emph{kernel function} $k:\mathcal{X}\times\mathcal{X}\longrightarrow\R$, $ K_{i,j} = k(\x_i,\x_j) = \varphi(\x_i)\T \varphi(\x_j), \; i\in[n], j\in[n],$ is the kernel matrix. We can now introduce the following proposition on the dual form of the previous optimization problem:

\begin{proposition}[Dual form of MC-SVM] Denote with $\alpha_{i,k}$ the $i-$th Lagrange multiplier relative to the $k-$th SVM, with $\bar{\alpha}_i = \Sum{k=1}{m}\alpha_{i,k}$ and define
\begin{equation}
\bar{\boldsymbol{\alpha}} =
\left[
\begin{array}{cccc}
\bar{\alpha}_1 &  \bar{\alpha}_2 & \cdots &  \bar{\alpha}_n 
\end{array}
\right]^\top.
\label{eq:alpha}
\end{equation}
The dual form of \eqref{eq: opt_prbl_primal} is 
\begin{eqnarray}
\label{eq:dual_form}
&\max\limits_{\boldsymbol{\bar{\alpha}}}& -\frac{\eta}{2}\bar{\boldsymbol{\alpha}}\T  D K D \bar{\boldsymbol{\alpha}}+ \boldsymbol{1}^\top\bar{\boldsymbol{\alpha}} \nonumber\\
& \text{\emph{s.t.}} & 
\bar{\alpha}_i = \Sum{k=1}{m} \alpha_{i,k}, \; i\in [n],\nonumber \\
&  & \Sum{i=1}{n}\alpha_{i,k}y_i=0, \; k\in [m],\\
&  & 0 \leq \alpha_{i,k} \leq  \frac{1}{2}\left(\left(1-2\tau_k\right)y_i+1\right), \nonumber \\
& & i \in [n], \  k\in[m]\nonumber.
\end{eqnarray}
The solution of \eqref{eq:dual_form} gives the equation for the separating hyperplane 
\begin{equation}
    \w = -\eta \Sum{i=1}{n} \bar{\alpha}_{i}y_i\varphi(\x_i).
    \label{eq:w}
\end{equation}
\label{prop: dual form}
\end{proposition}
\begin{figure*}[t!]
    \centering
    \begin{subfigure}[b]{0.25\textwidth}
        \includegraphics[width=\textwidth]{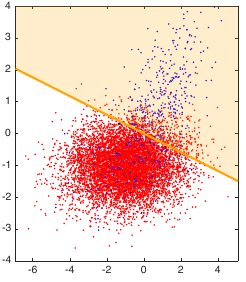}
        \captionsetup{justification=centering}
        \caption{Dataset $\mathcal{D}_{0.05}$, $\tau = 0.5$ (classic SVM).}
    \end{subfigure}
    \begin{subfigure}[b]{0.25\textwidth}
        \includegraphics[width=\textwidth]{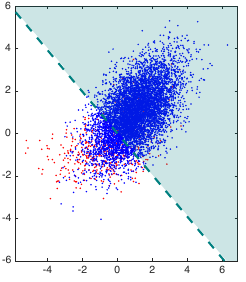}
        \captionsetup{justification=centering}
        \caption{Dataset $\mathcal{D}_{0.95}$, $\tau = 0.5$ (classic SVM).}
    \end{subfigure}
    \begin{subfigure}[b]{0.26\textwidth}
        \includegraphics[width=\textwidth]{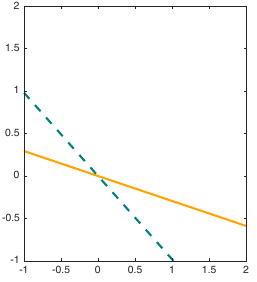}
        \captionsetup{justification=centering}
        \caption{The hyperplanes are very diferent.}
    \end{subfigure}
     \vfill
    \begin{subfigure}[b]{0.25\textwidth}
        \includegraphics[width=\textwidth]{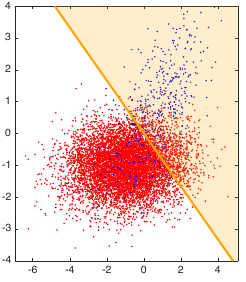}
        \captionsetup{justification=centering}
        \caption{Dataset $\mathcal{D}_{0.05}$, $\#\mathcal{T} = 5$.}
    \end{subfigure}
    \begin{subfigure}[b]{0.25\textwidth}
        \includegraphics[width=\textwidth]{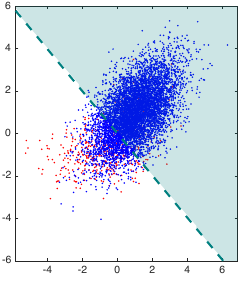}
        \captionsetup{justification=centering}
        \caption{Dataset $\mathcal{D}_{0.95}$, $\#\mathcal{T} = 5$.}
    \end{subfigure}
    \begin{subfigure}[b]{0.26\textwidth}
        \includegraphics[width=\textwidth]{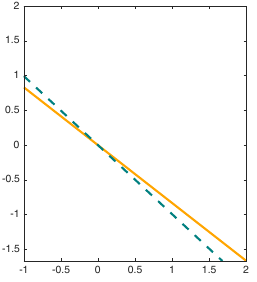}
        \captionsetup{justification=centering}
        \caption{The hyperplanes are closer. }
    \end{subfigure}
     \vfill
    \begin{subfigure}[b]{0.25\textwidth}
        \includegraphics[width=\textwidth]{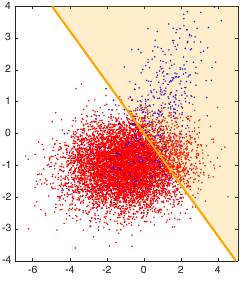}
        \captionsetup{justification=centering}
        \caption{Dataset $\mathcal{D}_{0.05}$, $\#\mathcal{T} = 10$.}
    \end{subfigure}
    \begin{subfigure}[b]{0.25\textwidth}
        \includegraphics[width=\textwidth]{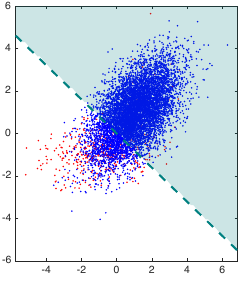}
        \captionsetup{justification=centering}
        \caption{Dataset $\mathcal{D}_{0.95}$, $\#\mathcal{T} = 10$}
    \end{subfigure}
    \begin{subfigure}[b]{0.26\textwidth}
        \includegraphics[width=\textwidth]{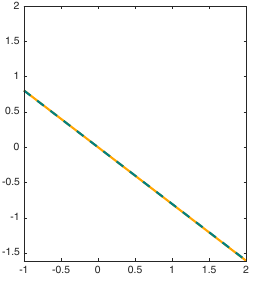}
        \captionsetup{justification=centering}
        \caption{The hyperplanes (almost) coincide.}
    \end{subfigure}
    \caption{Behavior of the classification hyperplane as $\tau$ varies. The more $\tau$ are used the more similar are the $\w$. The process ``saturates'' with $\#\mathcal{T} = 10$.}
    \label{fig:multipletaus}
\end{figure*}
The proof of Proposition \ref{prop: dual form} is given in the Appendix, Section \ref{proof_dualForm}. 

We notice that 
\begin{equation}
\begin{split}
\w\T 
\varphi(\x) =&  -\left(\eta \Sum{i=1}{n} \bar{\alpha}_{i}y_i\varphi(\x_i)\right)\T \varphi(\x) \\
=& -\eta \Sum{i=1}{n} \bar{\alpha}_{i}y_ik(\x_i,\x).
\end{split}
\label{eq:hyperplane}
\end{equation}
Thus, once $b$ has been chosen, a test point $\x$ would be classified as $S$ if 
$$ -\eta \Sum{i=1}{n} \bar{\alpha}_{i}y_ik(\x_i,\x)  - b < 0.$$
The choice of b is addressed in the next subsection (\ref{subsec: on the choice of b}).
\begin{remark}[Computational cost of MC-SVM]
As far as the computational cost is concerned, since the MC-SVM is closely related to the SVM, it can be estimated similarly to that of the SVM \cite{6796736}. In detail,  denoting by $n$ the number of points and by $d$ the number of features, the cost of SVM can be estimated as $O(\max(n,d)\cdot\min (n,d)^2)$. The only thing we have to consider is the complexity given by the weights stored in $\mathcal{T}$: being $m$ the number of weights used, the total cost of the MC-SVM will be $O(m\cdot(\max(n,d)\cdot\min (n,d)^2))$, since they appear as a summation in the minimization of the expected loss.
\end{remark}
%
\begin{example}
\label{ex:multiple_tau}
We set up an experiment to prove that the optimization problem described in \eqref{eq: opt_prbl_primal} leads to a classification boundary that is independent from the choice of $p_S$ and~$\varepsilon$.
We consider two Gaussian distributions corresponding to the safe and unsafe cases, with parameters: 
$$\boldsymbol{\mu}_S = [1,1]^\top, \Sigma_S = \begin{bmatrix}
1.5 & 0.8\\
0.8 & 1.5 
\end{bmatrix}; \quad
\boldsymbol{\mu}_U = [-1,-1]^\top, \Sigma_U = \begin{bmatrix}
2.5 & 0.1\\
0.1 & 0.5
\end{bmatrix}.
$$
Data set $\mathcal{D}_{0.05}$ is obtained by sampling with prior probability of safe points equal to $p_S = 0.05$  while data set $\mathcal{D}_{0.95}$ corresponds to sampling  with $p_s=0.95$. We then train four algorithms: two classical SVMs whose outputs are respectively $\w_{0.05}$ and $\w_{0.95}$ and two MC-SVMs whose outputs are $\tilde\w_{0.05}$ and $\tilde\w_{0.95}$. The idea is that $\w$s computed with classical SVM should be very different from each other and, in contrast, $\w$s computed with MC-SVM could be reasonably similar, meaning that the algorithm was able to learn the best separating hyperplane regardless of the sampling probability. The training set was built accordingly to the a priori probabilities of sampling for a total of $10000$ samples.
The kernel was linear and the regularization parameter $\eta$ was set to $10^{-3}$. 
For this experiment, three different configurations for $\mathcal{T}$ are used: a single value of $\tau$ equal to $0.5$ (classic SVM), and cases where $5$ and $10$ different values of $\tau$ are considered, obtained by sampling linearly spaced values between $0$ and $1$.\\
Figure \ref{fig:multipletaus} shows the behavior of the hyperplane classification. As expected, increasing the number of possible values of $\tau$ leads to a stabilization of the classifier.
\end{example}%
\subsection{On the choice of $b$}
\label{subsec: on the choice of b}
Once the form of the MC-SVM  (i.e. the weight vector $\w$) has been determined by solving \eqref{eq:dual_form}, we can exploit the degree of freedom provided by the bias parameter $b$ to ``adapt'' the classifier to the specific observed data. 

Formally, we assume to have a \textit{calibration set} of size $n_c$
\[
\mathcal{Z}_c\doteq\{(\tilde\x_i,\tilde y_i)\}_{i=1}^{n_c}\subseteq \mathcal{X}
\]
of new labelled data. Again, we define by $\mathcal{Z}_c^S$ (resp. $\mathcal{Z}_c^U$) the set of positive (resp. negative) data.
Note that this data will be associated to a particular safety probability $\tilde p_S$, and consequently, a different level of ``unbalance''. 

The design of $b$ is essential and it can be made in different ways. We revise some of the most suitable ones, at the best of our knowledge, before introducing a new method that takes into account both the confidence and the probability of observing the safe class.

\subsubsection{Bias adjustment} 

In the literature on unbalanced classification, it has been observed 
that the low presence of data in one class produces a shift of the separating hyperplane between the two classes towards the minority class, due to the request of reducing the total number of misclassifications. In extreme-cases of class imbalance, this would even lead to models labeling all the examples to the majority class \cite{Class_imbalance_survey}. In this case, solutions have been proposed to modify the hyperplane bias $b$ to adapt to this unbalance.

For instance, \cite{SVM_Bias} proposes to recompute the bias $b$ as
\begin{equation}
\label{eq:c_bias}
    b=\frac{b^U + b^S}{2}
\end{equation}
where 
\begin{align*}
b^U = \max_{\tilde{\x}_k \in \mathcal{Z}_c^U} \w^\top \tilde \x_k, \quad
b^S = \min_{\tilde{\x}_k \in \mathcal{Z}_c^S} \w^\top \tilde \x_k
\end{align*}
represent, respectively, the maximum value of the hyperplane without bias applied to the dataset of unsafe instances $\mathcal{Z}_c^U$, and the minimum value of the hyperplane without bias applied to the entries in the safe set $\mathcal{Z}_c^S$.

The definition of $b_s$ in \eqref{eq:c_bias}  has also been extended to take into account the
proportion of classes in the dataset, see again \cite{SVM_Bias,SVM_threshold_adjust}.
More advanced techniques, based on the same idea of shifting the SVM hyperplane, have been proposed in the literature.
For instance, in the z-SVM approach \cite{imam2006z},
the hyperplane is moved to a position such that the geometric mean of the accuracy of positive and negative
samples is maximized for the training data.
Clearly, the exact same approach can be translated into our setup.

\subsubsection{Adjustable classifier approach}

A second way to design the parameter $b$ stems from the realization that the MC-SVM is a \textit{adjustable classifier} according to the definition given in \cite{carlevaro2023probabilistic} (see in particular Assumption 1). Hence, we can exploit the solution proposed in that work. 

A (binary) classifier can be uniquely defined by its prediction function $\hat f(\x)$ that, depending on its sign, assigns labels $y$ to the sample $\x$. Such a function can be made \emph{adjustable} adding additively a scalar parameter $\rho$, i.e., $f(\x,\rho) = \hat f(\x) + \rho$. In this case, $\rho$ plays the role of the offset of the model. Clearly, in our case the role of the scaling parameter $\rho$ is played by $b$. Once the model is trained, it is easy to find the values $\bar\rho$ such that a chosen sample $\x$ lies on the boundary of the classifier by solving $f(\x,\bar\rho) = 0$.
Together with the calibration set $\mathcal{Z}_c$ consider $\delta\in (0,1)$, $\varepsilon\in (0,1)$, and an integer \textit{discarding parameter} $r$ satisfying $n_c\ge r \geq 1$ such that the Binomial cumulative distribution function be  
$$\mathbf{B}(r-1;n_c,\varepsilon)
\leq \delta.$$ 
Consider then the subset $\mathcal{Z}_c^{U}= \left\{(\tilde\x^U_j,-1)\right\}_{j=1}^{n_U} $ corresponding to all the negative samples in $\mathcal{Z}
_c$ and define the {\rm probabilistic scaling of level} $\varepsilon$ as follows
\begin{equation}
    \label{eq:rhoeps}
\bar\rho_\varepsilon \doteq
\mathrm{max}^{(r)}\left(\{\bar{\rho}(\tilde\x^U_{j})\}_{j=1}^{n_U}\right),
\end{equation}
where $\mathrm{max}^{(r)}$ denotes the $r-$\emph{maximum} of a list of values, i.e. such that there are no more than $r-1$ values greater than it. Defining then $\mathcal{S}(\rho) = \set{\x\in\mathcal{X} }{f(\x,\rho)<0}$, i.e. the set of samples that are predicted as $y=+1$ by the adjustable classifier,
we can eventually compute the $\bar\rho_\varepsilon$-\emph{safe set}
\begin{eqnarray*} 
\mathcal{S}_\varepsilon & \doteq & \bsis{cc} \mathcal{S}\left(\bar\rho_\varepsilon\right)  & \text{ if }  n_U\geq r \\ \mathcal{X} & \text{otherwise.} \esis
\end{eqnarray*} 
Then, with probability no smaller than $1-\delta$,  
\begin{equation}
\Pr\Bigl\{y=-1 \text{ and } \x \in \mathcal{S}_\varepsilon\Bigr\}  \leq \varepsilon.
\label{eq:theorem_Ps}
\end{equation}
Hence, we can adopt $\mathcal{S}_\varepsilon$ as a surrogate of $\Phi_\varepsilon$ and compute our bias as $b = \bar\rho_\varepsilon$.
\subsubsection{A-posteriori conformal prediction approach}

\emph{Conformal prediction} is another way to design the offset $b$. As stated in \cite{carlevaro2024conformal}, adjustable classifiers can be directly related to conformal prediction simply defining the following score function
$$s(\x,\hat y) = -\hat y\bar{\rho}(\x)$$
with $\bar{\rho}(\x)$ such that $f(\x,\bar\rho(\x))=0$. Hence, we can use conformal theory to design $b$. Indeed, the absolute value of the (almost) $(1-\varepsilon)-$quantile of the score values, $s_\varepsilon$, computed on a calibration set $\mathcal{Z}_c$ is proven to realize the same safety guarantee of \eqref{eq:theorem_Ps} but recent (still unpublished) research proved that the $(1-\varepsilon)-$quantile of the only ``negative'' score values $s_\varepsilon^U$ (i.e. the one computed on $\mathcal{Z}_c^U$) coincides exactly with $\bar\rho_\varepsilon$, ensuring the two methods bring to the same conclusions.

\subsubsection{An original approach for the design of $b$}

As commented after \eqref{eq:SVM_weighted}, when the regularization term $\eta$ is not considered, the weighted parameter $\tau$ can be interpreted as the false negative ratio, according to quantile regression theory \cite{koenker2005quantile}. In general, it is not possible to take small values of $\eta$, because this would hinder convergence of the SVM design. However, in the two-steps approach we propose here, we are designing $\w$ and $b$ independently.
Since we have proved that it is possible to construct a separating hyperplane that does not depend on the sampling probability (by solving \eqref{eq: opt_prbl_primal}), we solve a new optimization problem which does not depend on $\eta$ to find the best offset $b$ that realizes the desired amount of confidence in predicting false negatives. Given the ``original'' training set $\mathcal{Z}_T = \{(\x_i,y_i)\}_{i=1}^n$ (the one used to train \eqref{eq: opt_prbl_primal}), consider a calibration set $\mathcal{Z}_c$ obtained sampling \emph{with a chosen level of probability} $p_S$ from the same distribution of the whole data. The above considerations lead to the following optimization problem:
\begin{figure*}[h!]
     \centering
     \begin{subfigure}[b]{0.5\textwidth}
         \centering
         \includegraphics[width=\textwidth]{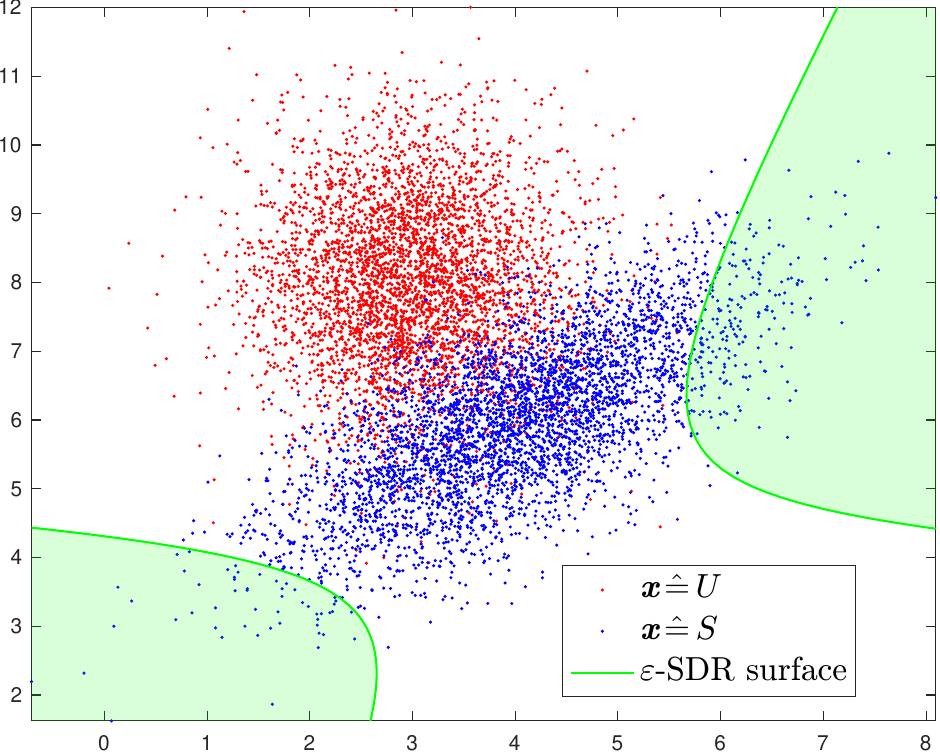}
         \caption{Exact Gaussian $\varepsilon$-SDR}
         \label{fig:exact}
     \end{subfigure}
     \vfill
     \begin{subfigure}[b]{0.32\textwidth}
         \centering
         \includegraphics[width=\textwidth]{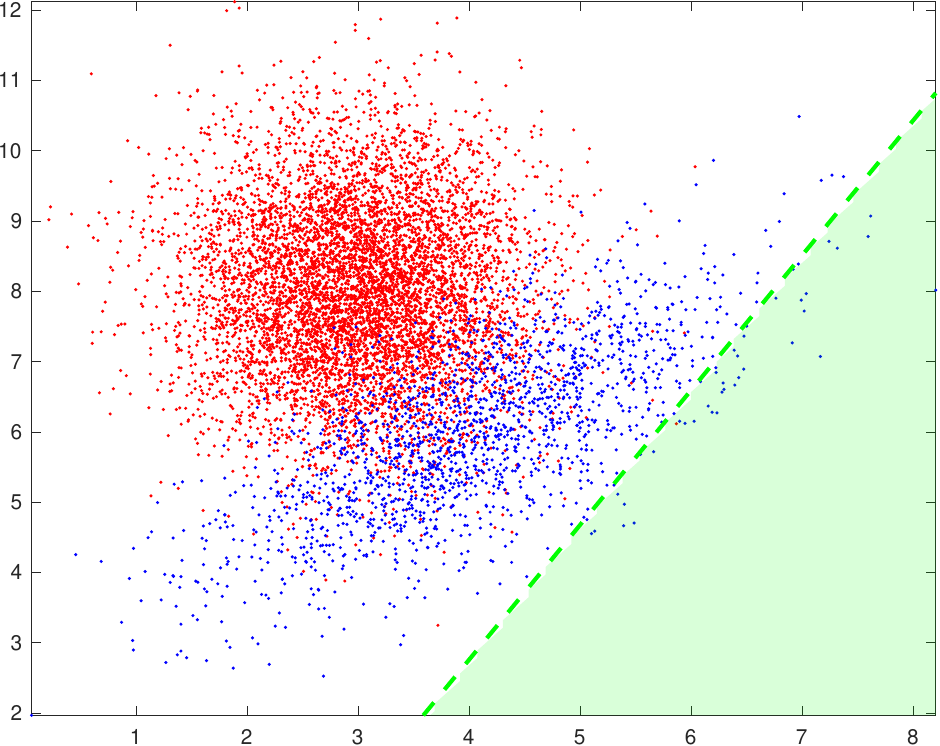}
         \caption{ Linear fit}
         \label{fig:one}
     \end{subfigure}
     \hfill
     \begin{subfigure}[b]{0.32\textwidth}
         \centering
         \includegraphics[width=\textwidth]{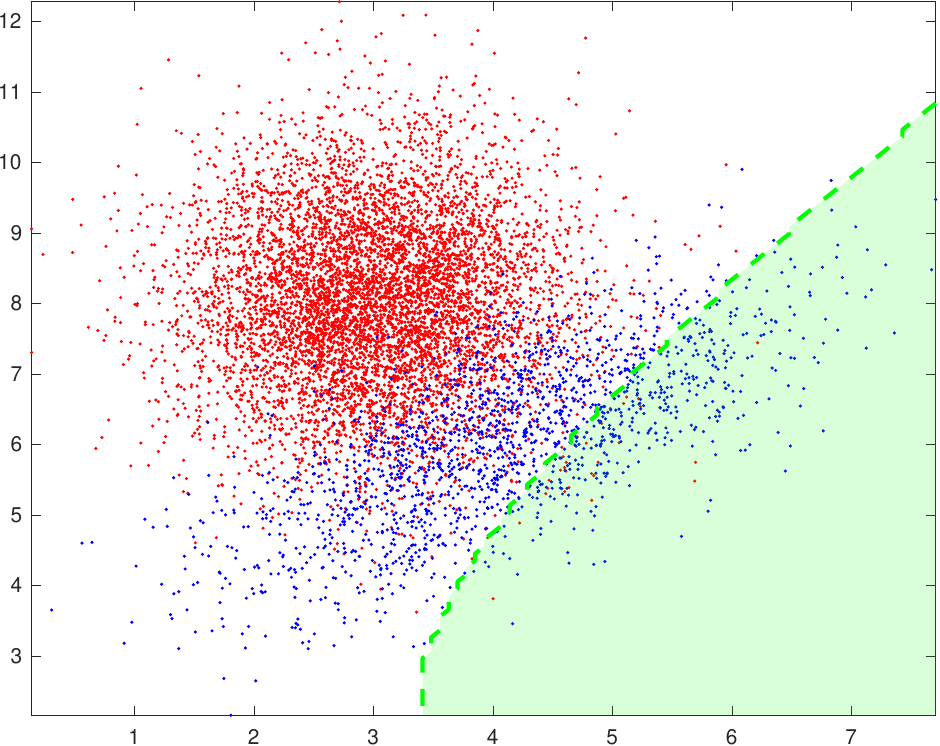}
         \caption{Quadratic fit}
         \label{fig:two}
     \end{subfigure}
     \hfill
     \begin{subfigure}[b]{0.32\textwidth}
         \centering
         \includegraphics[width=\textwidth]{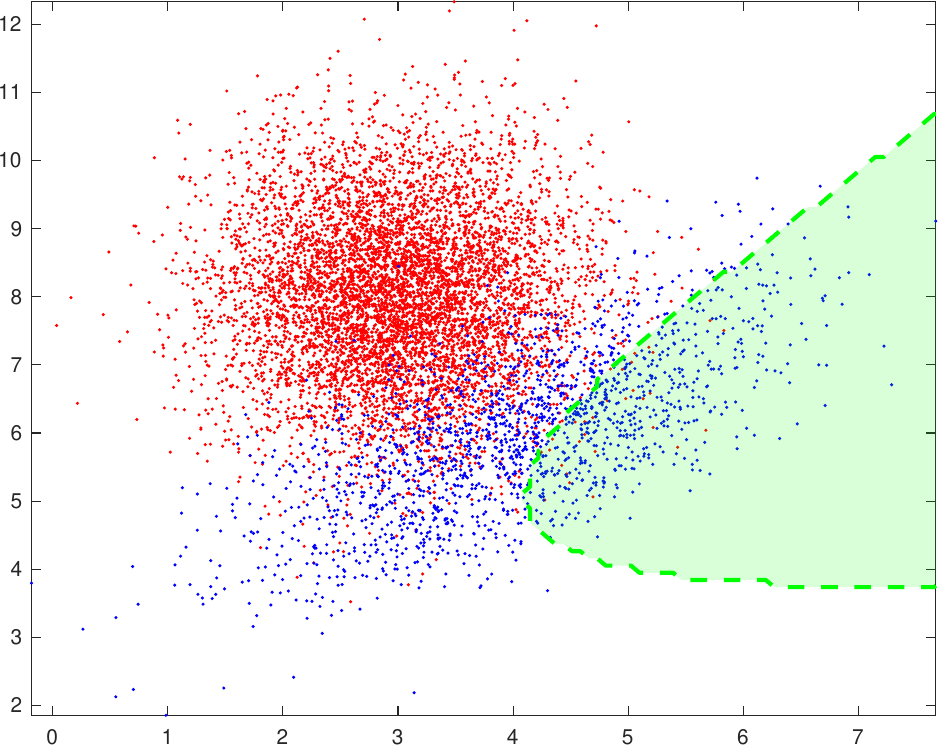}
         \caption{Cubic fit}
         \label{fig:five}
     \end{subfigure}
        \caption{Classification performance (for different kernels) after applying algorithm \eqref{eq: opt_prbl_primal} for the choice of an independent $\w$ and algorithm \eqref{eq:epsilonSVM} for the choice of $b$ such that the false positive ratio is bounded by $\varepsilon = 0.05$. Figure \ref{fig:exact} is the exact Gaussian $\varepsilon$-SDR.}
        \label{fig:approximation}
\end{figure*}
%
%
\begin{equation}
\min_{b} \; \frac{1}{2} \sum_{i=1}^{n_c} \max\left\{0, \left((1-2\varepsilon)\tilde y_i + 1\right)\left( \tilde y_i (\mathbf{w}^\top \varphi(\tilde\x_i) -b)+1\right)\right\}.
\label{eq:epsilonSVM}
\end{equation}
%
%
The solution of this problem (derived from \eqref{eq:SVM_weighted} neglecting the optimization of the hyperplane because already computed with \eqref{eq: opt_prbl_primal} and substituting the weight $\tau$ with~$\varepsilon$ because there is no longer any dependency on $\eta$) allows to design an offset $b$ that bounds, approximately, the number of false negatives in the prediction, maintaining them under $\varepsilon$. Of course, if one is interested in bounding the false positive ratio it is sufficient to substitute $\tau$ with $1-\varepsilon$. 
\begin{example}
In the same configuration as in Example \ref{example:PCRgaussian} we constructed a data set consisting of samples from 9 different probabilities of observing the ``Safe'' class ($y = +1$, the blue dots in the plots), equally sampled from $0.1$ to $0.9$ probabilities ($10000$ samples in total). For different kernels (linear, quadratic and cubic but more could be chosen), we trained the model \eqref{eq: opt_prbl_primal} as in Example \ref{ex:multiple_tau} with 9 different values of $\tau$, again equally spaced between $0.1$ to $0.9$. The regularization parameter $\eta$ was set to $10^{-3}$. As proved before, the vector of weights $\w$ obtained with such a procedure is independent on the choice of $p_S$, being the best hyperplane common to all the samples. Now, suppose that new data come, with a specific probability of sampling the ``Safe'' class $p_S' = 0.5$. Suppose further that the classification should not have more than $\varepsilon = 0.05$ false positives (``Unsafe'' points misclassified as ``Safe'' ones). To achieve this, we trained the model \eqref{eq:epsilonSVM} obtaining the special value $b$ that allows to classify the data with the required amount of confidence $1-\varepsilon = 0.95$ in the false positive ratio and for the predefined probability of sampling $p_S' = 0.5$. Specifically, the false positive rate for the exact $\varepsilon$-SDR and the linear, quadratic and cubic approximation are respectively: $0.009, 0, 0.002, 0.009$. Figure \ref{fig:approximation} shows the approximation of the exact Gaussian $\varepsilon$-SDR through the methodology described so far.
\end{example}
Following the example, it is worth noting that depending on the hyperparameters chosen for the classifier (kernel, regularization term, number of training points etc., etc.) the approximate $\varepsilon$-SDR $\tilde{\Phi}_\varepsilon$ is different. The point, however, is that with this procedure it is possible to build classifiers that maintain the desired property of a $\varepsilon$-SDR, i.e., that the classification boundary is independent on the choice of $p_S$ and $\varepsilon$ and, especially, that it is possible to bound the probability of observing an unsafe sample, i.e. with a sufficiently good approximation $\Pr\{U \; | \; \x\in \tilde{\Phi}_\varepsilon\}\le \varepsilon$. 

Before going to the conclusions, we still focus on the offset $b$ and the procedure we developed to design it. We started from a concept of radius (equation \eqref{eq:radius}) that depended on the probability $p_S$ and on the risk-level $\varepsilon$ and we retrieved (almost) the same dependency also here in the offset $b$. The fact that the problem \eqref{eq:epsilonSVM} is trained on a collection of data coming from the same probability of observing the class $S$ links the offset to $p_S$ and using $\varepsilon$ to control the number of false negatives (that, we recall, is possible because we have already computed~$\w$ and so there is no dependency on the regularization parameter $\eta$) links it to $\varepsilon$. So it is not incorrect to think about $b$ as a function of $p_S$ and $\varepsilon$, i.e. $b = b(p_S,\varepsilon)$. This, together with the relationship existing between $\w$ and $\Gamma(\x)$, traces a clear bridge between the first part of the paper on the exponential families of distributions and the approximated $\varepsilon$-Safe Decision Regions discussed in the second part.
\section{Conclusions}
Defining safety for data driven systems is an open problem in machine learning and this research investigates the question from a modeling perspective. First, the definition of $\varepsilon$-SDR (Definition
\ref{def: PCR}) is given to define formally what safety means in a classification problem (i.e. searching for the input features that realizes a declared safety event with high probability) and then it is specialized to the case of exponential distributions for which the $\varepsilon$-SDR takes the form of a closed set controlled by a radius that depends on the probability of observing the safe class $p_S$ and on the risk-level $\varepsilon$ (Prop. \ref{PropositionBella}). The request of exponentially distributed data is often not feasible in real applications, thus the remainder of the paper is dedicated to develop an SVM-based approximation of the $\varepsilon$-SDR that brings to the definition of \ref{eq: opt_prbl_primal}, an ensamble approach that can handle undersampled data or data coming from different probability of sampling (but always from the same distribution). Then, the confidence ($1-\varepsilon$) on the prediction is recovered exploiting the relationship between false positives/negatives and the SVM's weights (problem \ref{eq:epsilonSVM}). Thus, the contribution of this research has to be searched in the field of reliable AI and it is twofold: the definition of a ``safety framework'' for machine learning classification, the $\varepsilon$-SDR, and the implementation of \ref{eq: opt_prbl_primal}, a novel algorithm for heterogeneous classification. There are, however, also some limitations. The quality of the data is essential and the assumption of i.i.d. data (although  not specified) is necessary. And then \ref{eq: opt_prbl_primal} scales badly when the number of weights increase. Anyhow, these are two interesting starting points for future work on this topic alongside the implementation of the model for multi-class classification (a first, naive, idea could be exploiting ``one-vs-all'' approach on the target class). To conclude, this paper addresses a critical problem in modern machine learning, i.e. providing probabilistic guarantees of safety to the output of predictive models, enhancing the trustworthiness of data-driven classification models.

\acks{This work was supported in part by REXASI-PRO H-EU project, call HORIZON-CL4-2021-HUMAN-01-01, Grant agreement ID: 101070028. The work was also supported by Future Artificial Intelligence Research (FAIR) project, Italian National Recovery and Resilience Plan (PNRR), Spoke 3 - Resilient AI. The work of F. Dabbene was supported by the Italian Ministry of Research, under the complementary actions to the PNRR  “Fit4MedRob - Fit for Medical Robotics” Grant (PNC0000007). Moreover, T. Alamo acknowledges support from grants PID2022-142946NA-I00 and PID2022-141159OB-I00, funded by MICIU/AEI/ 10.13039/501100011033 and by ERDF/EU.}

\appendix
\section{Characterization of the $\varepsilon$-SDR for Gaussian distribution}
\label{appendix:PCRgaussian}
Focusing on a specific exponential family, the Gaussian distribution, we can give an exact description of the $\varepsilon$-SDR.
Assume  that $f(\x|S)$ and $f(\x|U)$ are Gaussian, that is:
\begin{eqnarray*} f(\x|S) &=& \dfrac{1}{c_S} \exp\left(-\frac{1}{2}(\x-{\boldsymbol{\mu}_S})^\top \Sigma^{-1}_S (\x-{\boldsymbol{\mu}_S})\right), \\
f(\x|U) &=& \dfrac{1}{c_U} \exp\left(-\frac{1}{2}(\x-\boldsymbol{\mu}_U)^\top \Sigma^{-1}_U(\x-\boldsymbol{\mu}_U)\right)
\end{eqnarray*}

where, respectively for $S$ and $U$,
$c_S = \sqrt{\det(2\pi \Sigma_S)}$
and $c_U = \sqrt{\det(2\pi \Sigma_U)}$,  $\boldsymbol{\mu}_S, \boldsymbol{\mu}_U$ are the means and $\Sigma_S, \Sigma_U$ are the covariance matrices. 

This corresponds to 
\begin{eqnarray*}
g_S(\x) &=& \frac{1}{2}(\x-\boldsymbol{\mu}_S)^\top \Sigma^{-1}_S(\x-\boldsymbol{\mu}_S), \\
g_U(\x) &=& \frac{1}{2}(\x-\boldsymbol{\mu}_U)^\top \Sigma^{-1}_U(\x-\boldsymbol{\mu}_U), 
\end{eqnarray*} 
By analyzing the difference between the exponential components of the distributions, it is possible to better define the form of the $\varepsilon$-SDR:
\begin{eqnarray*}
    g_S(\x) - g_U(\x) &=& \frac{1}{2}\left((\x-\boldsymbol{\mu}_S)^\top \Sigma^{-1}_S(\x-\boldsymbol{\mu}_S) - (\x-\boldsymbol{\mu}_U)^\top \Sigma^{-1}_U(\x-\boldsymbol{\mu}_U) \right) \\
    &=& \frac{1}{2}\bigl(\x^\top\left(\Sigma_S^{-1} - \Sigma_U^{-1}\right)\x - 2\x^\top\left(\Sigma_S^{-1}\boldsymbol{\mu}_S-\Sigma_U^{-1}\boldsymbol{\mu}_U\right) + \boldsymbol{\mu}_S^\top \Sigma^{-1}_S\boldsymbol{\mu}_S  \\ 
    & & -\boldsymbol{\mu}_U^\top \Sigma^{-1}_U\boldsymbol{\mu}_U \bigr).
\end{eqnarray*}
Thus, the behavior of the $\varepsilon$-SDR is uniquely defined by the difference between the inverses of the covariance matrices, i.e., $\mathrm{A} = \Sigma_S^{-1}-\Sigma_U^{-1}$. Let us study it in detail: 
\paragraph{Case 1. $\Sigma_S \textrm{ and } \Sigma_U$ are  scalar multiples, i.e. 
$\Sigma_U = k\Sigma_S$, $k\in\R\setminus\{0\}$:}
\begin{itemize}
    \item $k=1$.\\ 
    In this case the $\varepsilon$-SDR is an hyperplane:
    \begin{equation*}
    \Phi_\varepsilon = \set{\x}{\boldsymbol{a}^\top\x  + \boldsymbol{b} \le \rho\left(p_S,\varepsilon\right)}, 
    \end{equation*}  
    where
    \begin{eqnarray*}
     \boldsymbol{a} &=& -\Sigma_S^{-1}\left(\boldsymbol{\mu}_S-\boldsymbol{\mu}_U\right), \\
    \boldsymbol{b} &=& \frac{1}{2}\left(\boldsymbol{\mu}_S^\top \Sigma^{-1}_S\boldsymbol{\mu}_S - \boldsymbol{\mu}_U^\top \Sigma^{-1}_S\boldsymbol{\mu}_U\right) + \ln\frac{c_S}{c_U}.
    \end{eqnarray*}
    \item $k\neq 1$.\\
    In this case the $\varepsilon$-SDR is an ellipsoid (positive or negative oriented):
    \begin{equation*}
    \Phi_\varepsilon = \set{\x}{\frac{1}{2}(\x-\boldsymbol{\mu}_k)^\top \Sigma_k^{-1}(\x-\boldsymbol{\mu}_k) + \gamma_k \le \rho\left(p_S,\varepsilon\right)},
    \end{equation*}  
    where
     \begin{eqnarray*}
     \Sigma_k^{-1} &=& \left(1-\frac{1}{k}\right)\Sigma_S^{-1},\\ 
     \boldsymbol{\mu}_k &=&\frac{1}{k-1}\left(k\boldsymbol{\mu}_S-\boldsymbol{\mu}_U\right),\\
     \gamma_k &=& -\boldsymbol{\mu}_k^\top \Sigma^{-1}_k\boldsymbol{\mu}_k +\boldsymbol{\mu}_S^\top \Sigma^{-1}_S\boldsymbol{\mu}_S -\boldsymbol{\mu}_U^\top \frac{1}{k}\Sigma^{-1}_S\boldsymbol{\mu}_U + \ln\frac{c_S}{c_U}.
    \end{eqnarray*}
    The orientation of the ellipsoid is negative if $k\in(0,1)$ and positive otherwise.
\end{itemize}
\paragraph{Case 2.
$\Sigma_S \textrm{ and } \Sigma_U$ are not scalar multiples:} 
  In this case the $\varepsilon$-SDR is a quadric: 
    \begin{equation*}
    \begin{split}
    \Phi_\varepsilon = \set{\x}{\frac{1}{2}\x^\top \mathrm{A}\x +\boldsymbol{a}^\top\x + \boldsymbol{b} \le \rho\left(p_S,\varepsilon\right)}, 
    \end{split}
    \end{equation*}
where
\begin{eqnarray*}
    \boldsymbol{a} &=& -\left(\Sigma^{-1}_S\boldsymbol{\mu}_S-\Sigma^{-1}_U\boldsymbol{\mu}_U\right),\\
    \boldsymbol{b} &=& \boldsymbol{\mu}_S^\top \Sigma^{-1}_S\boldsymbol{\mu}_S -\boldsymbol{\mu}_U^\top \Sigma^{-1}_U\boldsymbol{\mu}_U + \ln\frac{c_S}{c_U}.
\end{eqnarray*}
In particular, when $\mathrm{A}$ is invertible we can write it in a more compact form:
\begin{equation*}
\begin{split}
\Phi_\varepsilon = \set{\x}{\frac{1}{2}(\x-\boldsymbol{\mu})^\top \mathrm{A}(\x-\boldsymbol{\mu}) + \gamma' \le \rho\left(p_S,\varepsilon\right)}, 
\end{split}
\end{equation*}
where
\begin{eqnarray*}
 \boldsymbol{\mu} &=& \mathrm{A}^{-1}\left(\Sigma^{-1}_S\boldsymbol{\mu}_S-\Sigma^{-1}_U\boldsymbol{\mu}_U\right), \\
\gamma' &=& -\boldsymbol{\mu}^\top \Sigma\boldsymbol{\mu} +\boldsymbol{\mu}_S^\top \Sigma^{-1}_S\boldsymbol{\mu}_S -\boldsymbol{\mu}_U^\top \Sigma^{-1}_U\boldsymbol{\mu}_U + \ln\frac{c_S}{c_U}.
\end{eqnarray*}
\section{Proofs}

\subsection{Proof of Preposition \ref{PropositionBella}}
\label{Proof: PropostionBella}

\begin{proof}
It is clear that, by definition, $\x\in \Phi_\varepsilon$ implies that event $S$ occurs for $\x$ with probability at least $1-\varepsilon$. 
Thus, 
$$ p(U|\x) = 1- p(S|\x) \leq \varepsilon, \; \forall \x\in \Phi_\varepsilon.$$
Given $p_S\in (0,1)$, $\varepsilon \in (0,1)$ and the density function $f(\x|S)$, we can compute the $\varepsilon$-SDR  from 
\begin{equation}\label{equ:pAx}
    p(S|\x)f(\x) = f(\x|S) p_S.
\end{equation} 
Thus, 
$$ p(S|\x)= \frac{ f(\x|S) p_S}{f(\x)} = \frac{f(\x|S) p_S}{f(\x|S) p_S+f(\x|U) p_U}.$$
Since $\dfrac{z}{1-z}$ is monotonically growing in $[0,1)$,
\begin{equation}
p(S|\x) \ge 1-\varepsilon \iff \frac{p(S|\x)}{1-p(S|\x)} \ge \frac{1-\varepsilon}{\varepsilon}.
\label{ineq:on:p:A:x}
\end{equation}
\noindent From \eqref{equ:pAx} we have 
$$ p(S|\x) = \frac{f(\x|S)p_S}{f(\x)} .$$  
Similarly, 
$$ p(U|\x) = \frac{f(\x|U)p_U}{f(\x)} .$$  
\noindent Thus, 
\noindent we get
\begin{equation*}
\frac{p(S|\x)}{1-p(S|\x)} = \frac{p(S|\x)}{p(U|\x)} = \frac{f(\x|S)p_S}{f(\x|U)p_U}.
\end{equation*}
This, along with \eqref{ineq:on:p:A:x}, yields
$$ p(S|\x) \ge 1-\varepsilon \iff \frac{f(\x|S)p_S}{f(\x|U)p_U} \ge \frac{1-\varepsilon}{\varepsilon}. $$
\noindent Then,
\begin{eqnarray*}
    \Phi_\varepsilon &=& \set{\x}{p(S|\x) \ge 1-\varepsilon} \\
    &=& \set{\x}{f(\x|S) \ge \left(\dfrac{1-\varepsilon}{\varepsilon}\right)\dfrac{p_U}{p_S}f(\x|U)}\\
    &=& \set{\x}{f(\x|S) \ge \left(\dfrac{1-\varepsilon}{\varepsilon}\right)\left(\dfrac{1-p_S}{p_S}\right)f(\x|U)}.
\end{eqnarray*}
Taking logarithms and using $ \rho(p_S,\varepsilon) = \ln\left(\dfrac{\varepsilon}{1-\varepsilon}\dfrac{p_S}{1-p_S}\right) $, we obtain
\begin{eqnarray*}
     \Phi_\varepsilon  &=& \{ \; \x \; : \;  \ln (f(x|S)) \geq \rho(p_S,\varepsilon) + \ln (f(x|U))\; \} \\
     & = & \{\; \x \; : \; -g_S(\x) - \ln\,c_S   \ge -\rho(p_S,\varepsilon) -g_U(\x) - \ln\, c_U \; \} \\
       &=& \set{\x}{\tilde{g}_S(\x) - \tilde{g}_U(\x) \le \rho(p_S,\varepsilon)}\\
       &=& \set{\x}{\Gamma(\x)\le\rho(p_S,\varepsilon)}.
\end{eqnarray*}

\end{proof}

\subsection{Proof of Preposition \ref{prop: dual form}}
\begin{proof}
\label{proof_dualForm}
Defining the multipliers
$$ \boldsymbol{\alpha}_i= \bv \alpha_{i,1} \\ \alpha_{i,2} \\ \vdots \\ \alpha_{i,m} \ev \; \; \boldsymbol{\beta}_i= \bv \beta_{i,1} \\ \beta_{i,2} \\ \vdots \\ \beta_{i,m}\ev, \; i\in [n],$$
with $\alpha_{i,k}, \beta_{i,k}\ge0$ for $i\in[n], k\in[m]$, the Lagrangian is given by
\begin{eqnarray*}
\mathcal{L}&=&  \mathcal{L}(\w,\mathbf{b},\boldsymbol{\xi}_1,\dots, \xiB_n,\boldsymbol{\alpha}_1,\ldots,\boldsymbol{\alpha}_n,\boldsymbol{\beta}_1,\ldots,\boldsymbol{\beta}_n) \\
&=&\frac{1}{2\eta}\w^\top\w + \frac{1}{2} \Sum{k=1 }{m} \Sum{i=1 }{n} \left((1-2\tau_k)y_i +1\right) \xi_{i,k} + \\
& &\Sum{i=1}{n}\sum_{k=1}^m \left(\alpha_{i,k}\left(y_i(\w^\top \varphi(\x_i) -b_k)+1-\xi_{i,k}\right)  - \beta_{i,k}\xi_{i,k}\right)\\
&=&\frac{1}{2\eta}\w^\top\w + \frac{1}{2} \Sum{k=1 }{m} \Sum{i=1 }{n} \left((1-2\tau_k)y_i +1\right) \xi_{i,k} + \\
& &\Sum{i=1}{n}\left(\sum_{k=1}^m \alpha_{i,k} \right) y_i \w^\top \varphi(\x_i)+\Sum{i=1}{n}\sum_{k=1}^m \left( \alpha_{i,k}\left( -y_i b_k +1 -\xi_{i,k}\right)-\beta_{i,k}\xi_{i,k} \right).
\end{eqnarray*}
Let us define 
$$\bar{\alpha}_i= \Sum{k=1}{m}\alpha_{i,k}.$$
By substitution and rearranging some terms, we obtain
\begin{eqnarray}
\label{eq: lagr 1}
\mathcal{L} &=& \frac{1}{2\eta}\w^\top\w + \nonumber\\  
& &\frac{1}{2} \Sum{k=1 }{m} \Sum{i=1 }{n} \left((1-2\tau_k)y_i +1  -2\alpha_{i,k} - 2\beta_{i,k}\right)  \xi_{i,k} + \\
& &\left( \Sum{i=1}{n}\bar{\alpha}_i y_i \varphi(x_i) \right)^\top\w  + \Sum{i=1}{n} \bar{\alpha}_i-\Sum{i=1}{n}\sum_{k=1}^m \alpha_{i,k}  y_i b_k. \nonumber
\end{eqnarray}
With this form for the Lagrangian, setting partial derivatives to zero gives the following constraints:
\begin{eqnarray}
    \parder{\mathcal{L}}{\w} &=& 0 \Rightarrow \w = -\eta \Sum{i=1}{n} \bar{\alpha}_{i}y_i\varphi(\x_i)
    \label{eq: w}\\
    \parder{\mathcal{L}}{b_k} &=& 0 \Rightarrow \sum_{i=1}^n\alpha_{i,k} y_i=0,\; k \in [m] 
    \label{eq: dbk=0}\\
    \parder{\mathcal{L}}{\xi_{i,k}} &=& 0 \Rightarrow \beta_{i,k} = -\alpha_{i,k}+\frac{1}{2}\left(\left(1-2\tau_k\right)y_i+1\right) 
    \nonumber\\
    &\Rightarrow& 0 \le \alpha_{i,k} \le \frac{1}{2}\left(\left(1-2\tau_k\right)y_i+1\right), \quad i \in [n], \  k\in[m] ,
    \label{eq: dxi=0}
\end{eqnarray}
with \eqref{eq: dxi=0} obtained exploiting the non-negativeness of the multipliers. 
Substituting equations (\ref{eq: w} - \ref{eq: dxi=0}) into (\ref{eq: lagr 1}) we obtain:
\begin{eqnarray}
    \label{eq:lag2}
    \mathcal{L} &=& \mathcal{L}(\boldsymbol{\alpha}_1,\ldots,\boldsymbol{\alpha}_n,\bar\alpha_1, \dots, \bar\alpha_n) \nonumber \\
    &=& -\frac{\eta}{2}\left(\Sum{i=1}{n}y_i \bar{\alpha}_i \varphi(\x_i)\right)^\top \left(\Sum{i=1}{n}y_i \bar{\alpha}_i \varphi(\x_i)\right) + \Sum{i=1}{n} \bar{\alpha}_i.
\end{eqnarray}
The objective is now maximizing $\mathcal{L}$ with respect to the dual variables $\alpha_{i,k}$ and  $\bar{\alpha}_i$, $i\in[n]$, $k\in [m]$, under the new constraints:
\begin{eqnarray*}
\bar{\alpha}_i & = & \Sum{k=1}{m} \alpha_{i,k}, \; i\in [n], \\
0 & = & \Sum{i=1}{n} \alpha_{i,k} y_i, \; k\in [m],\\
0 & \leq & \alpha_{i,k} \leq  \frac{1}{2}\left(\left(1-2\tau_k\right)y_i+1\right), \;  i \in [n], \  k\in[m].
\end{eqnarray*}
For a more compact notation, we use \eqref{eq:R}, \eqref{eq:D}, \eqref{eq:K} and \eqref{eq:alpha} 
%
%
to write the Lagrangian $\mathcal{L}$ in this form
%
%
%
%
\begin{equation}
\mathcal{L} = -\frac{\eta}{2}\bar{\boldsymbol{\alpha}}\T  D K D \bar{\boldsymbol{\alpha}}+ \boldsymbol{1}^\top\bar{\boldsymbol{\alpha}}. 
\end{equation}
\end{proof}

\vskip 0.2in
\bibliography{references.bib}

\begin{thebibliography}{32}
\providecommand{\natexlab}[1]{#1}
\providecommand{\url}[1]{\texttt{#1}}
\expandafter\ifx\csname urlstyle\endcsname\relax
  \providecommand{\doi}[1]{doi: #1}\else
  \providecommand{\doi}{doi: \begingroup \urlstyle{rm}\Url}\fi

\bibitem[Abramovich and Ritov(2013)]{abramovich2013statistical}
F.~Abramovich and Y.~Ritov.
\newblock \emph{Statistical Theory: A Concise Introduction}.
\newblock Chapman and Hall/CRC, 1st edition, 2013.
\newblock \doi{10.1201/b14755}.

\bibitem[Angelopoulos and Bates(2023)]{angelopoulos2023conformal}
Anastasios~N. Angelopoulos and Stephen Bates.
\newblock Conformal prediction: A gentle introduction.
\newblock \emph{Foundations and Trends® in Machine Learning}, 16\penalty0 (4):\penalty0 494--591, 2023.
\newblock ISSN 1935-8237.
\newblock \doi{10.1561/2200000101}.

\bibitem[Cao et~al.(2013)Cao, Zhao, and Zaïane]{CS-SVM}
P.~Cao, Dazhe Zhao, and Osmar Zaïane.
\newblock An optimized cost-sensitive svm for imbalanced data learning.
\newblock In \emph{Advances in Knowledge Discovery and Data Mining}, pages 280--292, 04 2013.
\newblock ISBN 978-3-642-37455-5.
\newblock \doi{10.1007/978-3-642-37456-2_24}.

\bibitem[Carlevaro and Mongelli(2022)]{9594676}
Alberto Carlevaro and Maurizio Mongelli.
\newblock \protect{A New SVDD Approach to Reliable and Explainable AI}.
\newblock \emph{IEEE Intelligent Systems}, 37\penalty0 (2):\penalty0 55--68, 2022.
\newblock \doi{10.1109/MIS.2021.3123669}.

\bibitem[Carlevaro et~al.(2023)Carlevaro, Alamo, Dabbene, and Mongelli]{carlevaro2023probabilistic}
Alberto Carlevaro, Teodoro Alamo, Fabrizio Dabbene, and Maurizio Mongelli.
\newblock Probabilistic safety regions via finite families of scalable classifiers, 2023.

\bibitem[Carlevaro et~al.(2024{\natexlab{a}})Carlevaro, Alamo, Dabbene, and Mongelli]{carlevaro2024conformal}
Alberto Carlevaro, Teodoro Alamo, Fabrizio Dabbene, and Maurizio Mongelli.
\newblock \protect{Conformal predictions for probabilistically robust scalable machine learning classification}.
\newblock \emph{Machine Learning}, pages 1--17, 2024{\natexlab{a}}.
\newblock \doi{https://doi.org/10.1007/s10994-024-06571-6}.

\bibitem[Carlevaro et~al.(2024{\natexlab{b}})Carlevaro, Narteni, Dabbene, Alamo, and Mongelli]{carlevaroCOPA}
Alberto Carlevaro, Sara Narteni, Fabrizio Dabbene, Teodoro Alamo, and Maurizio Mongelli.
\newblock A probabilistic scaling approach to conformal predictions in binary image classification.
\newblock In \emph{Conformal and Probabilistic Prediction with Applications}. PMLR, 2024{\natexlab{b}}.

\bibitem[Chapelle(2007)]{6796736}
Olivier Chapelle.
\newblock Training a support vector machine in the primal.
\newblock \emph{Neural Computation}, 19\penalty0 (5):\penalty0 1155--1178, 2007.
\newblock \doi{10.1162/neco.2007.19.5.1155}.

\bibitem[Cortes and Vapnik(1995)]{SVM}
Corinna Cortes and Vladimir Vapnik.
\newblock Support-vector networks.
\newblock \emph{Mach. Learn.}, 20\penalty0 (3):\penalty0 273–297, sep 1995.
\newblock \doi{10.1023/A:1022627411411}.

\bibitem[Damiani and Ardagna(2020)]{damiani2020certified}
Ernesto Damiani and Claudio~A. Ardagna.
\newblock Certified machine-learning models.
\newblock In \emph{46th International Conference on Current Trends in Theory and Practice of Informatics, SOFSEM 2020, Limassol, Cyprus, January 20–24, 2020, Proceedings}, page 3–15, Berlin, Heidelberg, 2020. Springer-Verlag.
\newblock \doi{10.1007/978-3-030-38919-2_1}.

\bibitem[Dong et~al.(2020)Dong, Yu, Cao, Shi, and Ma]{dong2020survey}
Xibin Dong, Zhiwen Yu, Wenming Cao, Yifan Shi, and Qianli Ma.
\newblock A survey on ensemble learning.
\newblock \emph{Frontiers of Computer Science}, 14\penalty0 (2):\penalty0 241, 2020.
\newblock \doi{10.1007/s11704-019-8208-z}.

\bibitem[He and Ma(2013)]{Class_imbalance_survey}
Haibo He and Yunqian Ma.
\newblock \emph{Class Imbalance Learning Methods for Support Vector Machines}, pages 83--99.
\newblock 2013.
\newblock \doi{10.1002/9781118646106.ch5}.

\bibitem[Imam et~al.(2006)Imam, Ting, and Kamruzzaman]{imam2006z}
Tasadduq Imam, Kai~Ming Ting, and Joarder Kamruzzaman.
\newblock \protect{z-SVM: An SVM for improved classification of imbalanced data}.
\newblock In \emph{AI 2006: Advances in Artificial Intelligence: 19th Australian Joint Conference on Artificial Intelligence, Hobart, Australia, December 4-8, 2006. Proceedings 19}, pages 264--273. Springer, 2006.
\newblock \doi{10.1007/11941439_30}.

\bibitem[Kalid et~al.(2020)Kalid, Ng, Tong, and Khor]{kalid2020multiple}
Suraya~Nurain Kalid, Keng-Hoong Ng, Gee-Kok Tong, and Kok-Chin Khor.
\newblock \protect{A Multiple Classifiers System for Anomaly Detection in Credit Card Data With Unbalanced and Overlapped Classes}.
\newblock \emph{IEEE Access}, 8:\penalty0 28210--28221, 2020.
\newblock \doi{10.1109/ACCESS.2020.2972009}.

\bibitem[Kaur et~al.(2022)Kaur, Uslu, Rittichier, and Durresi]{kaur2022trustworthy}
Davinder Kaur, Suleyman Uslu, Kaley~J. Rittichier, and Arjan Durresi.
\newblock \protect{Trustworthy Artificial Intelligence: A Review}.
\newblock \emph{ACM Comput. Surv.}, 55\penalty0 (2), jan 2022.
\newblock ISSN 0360-0300.
\newblock \doi{10.1145/3491209}.

\bibitem[Koenker(2005)]{koenker2005quantile}
Roger Koenker.
\newblock \emph{Quantile Regression}, volume~38 of \emph{Econometric Society Monographs}.
\newblock Cambridge University Press, Cambridge, July 2005.
\newblock \doi{10.1017/CBO9780511754098}.

\bibitem[Li et~al.(2023)Li, Qi, Liu, Di, Liu, Pei, Yi, and Zhou]{10.1145/3555803}
Bo~Li, Peng Qi, Bo~Liu, Shuai Di, Jingen Liu, Jiquan Pei, Jinfeng Yi, and Bowen Zhou.
\newblock \protect{Trustworthy AI: From Principles to Practices}.
\newblock \emph{ACM Comput. Surv.}, 55\penalty0 (9), jan 2023.
\newblock ISSN 0360-0300.
\newblock \doi{10.1145/3555803}.

\bibitem[Liu et~al.(2022)Liu, Wang, Fan, Liu, Li, Jain, Liu, Jain, and Tang]{liu2022trustworthy}
Haochen Liu, Yiqi Wang, Wenqi Fan, Xiaorui Liu, Yaxin Li, Shaili Jain, Yunhao Liu, Anil Jain, and Jiliang Tang.
\newblock \protect{Trustworthy AI: A Computational Perspective}.
\newblock \emph{ACM Transactions on Intelligent Systems and Technology}, 14, 07 2022.
\newblock \doi{10.1145/3546872}.

\bibitem[Maleki et~al.(2020)Maleki, Muthukrishnan, Ovens, Md, and Forghani]{maleki2020machine}
Farhad Maleki, Nikesh Muthukrishnan, Katie Ovens, Caroline Md, and Reza Forghani.
\newblock \protect{Machine Learning Algorithm Validation: From Essentials to Advanced Applications and Implications for Regulatory Certification and Deployment}.
\newblock \emph{Neuroimaging Clinics of North America}, 30:\penalty0 433--445, 11 2020.
\newblock \doi{10.1016/j.nic.2020.08.004}.

\bibitem[Menzies and Pecheur(2005)]{menzies2005verification}
Tim Menzies and Charles Pecheur.
\newblock \protect{Verification and Validation and Artificial Intelligence}.
\newblock volume~65 of \emph{Advances in Computers}, pages 153--201. Elsevier, 2005.
\newblock \doi{10.1016/S0065-2458(05)65004-8}.

\bibitem[Muhammad et~al.(2021)Muhammad, Ullah, Lloret, Ser, and {De Albuquerque}]{muhammad2020deep}
Khan Muhammad, Amin Ullah, Jaime Lloret, {Javier Del} Ser, and {Victor Hugo C.} {De Albuquerque}.
\newblock \protect{Deep Learning for Safe Autonomous Driving: Current Challenges and Future Directions}.
\newblock \emph{IEEE Transactions on Intelligent Transportation Systems}, 22\penalty0 (7):\penalty0 4316--4336, July 2021.
\newblock ISSN 1524-9050.
\newblock \doi{10.1109/TITS.2020.3032227}.
\newblock Publisher Copyright: {\textcopyright} 2000-2011 IEEE.

\bibitem[Myllyaho et~al.(2021)Myllyaho, Raatikainen, M{\"a}nnist{\"o}, Mikkonen, and Nurminen]{myllyaho2021systematic}
{Lalli Santeri} Myllyaho, Mikko Raatikainen, Tomi M{\"a}nnist{\"o}, Tommi Mikkonen, and {Jukka K} Nurminen.
\newblock \protect{Systematic literature review of validation methods for AI systems}.
\newblock \emph{The Journal of Systems and Software}, 181, November 2021.
\newblock ISSN 0164-1212.
\newblock \doi{10.1016/j.jss.2021.111050}.

\bibitem[N{\'u}{\~n}ez et~al.(2017)N{\'u}{\~n}ez, Gonzalez-Abril, and Angulo]{SVM_Bias}
Haydemar N{\'u}{\~n}ez, Luis Gonzalez-Abril, and Cecilio Angulo.
\newblock \protect{Improving SVM Classification on Imbalanced Datasets by Introducing a New Bias}.
\newblock \emph{Journal of Classification}, 34\penalty0 (3):\penalty0 427--443, 2017.
\newblock \doi{10.1007/s00357-017-9242-x}.

\bibitem[Pietraszek and Tanner(2005)]{PIETRASZEK2005169}
Tadeusz Pietraszek and Axel Tanner.
\newblock \protect{Data mining and machine learning—Towards reducing false positives in intrusion detection}.
\newblock \emph{Information Security Technical Report}, 10\penalty0 (3):\penalty0 169--183, 2005.
\newblock ISSN 1363-4127.
\newblock \doi{10.1016/j.istr.2005.07.001}.

\bibitem[Schaefer et~al.(2020)Schaefer, Lehne, Schepers, Prasser, and Thun]{schaefer2020use}
Julia Schaefer, Moritz Lehne, Josef Schepers, Fabian Prasser, and Sylvia Thun.
\newblock \protect{The use of machine learning in rare diseases: A scoping review}.
\newblock \emph{Orphanet Journal of Rare Diseases}, 15:\penalty0 1--10, 12 2020.
\newblock \doi{10.1186/s13023-020-01424-6}.

\bibitem[Seto and Sha(1999)]{SetoAnEngineering1999}
Danbing Seto and Lui Sha.
\newblock An engineering method for safety region development.
\newblock Technical Report CMU/SEI-99-TR-018, Software Engineering Institute, Carnegie Mellon University, Pittsburgh, PA, 1999.

\bibitem[Shanahan and Roma(2003)]{SVM_threshold_adjust}
James~G. Shanahan and Norbert Roma.
\newblock \protect{Improving SVM Text Classification Performance through Threshold Adjustment}.
\newblock In Nada Lavra{\v{c}}, Dragan Gamberger, Hendrik Blockeel, and Ljup{\v{c}}o Todorovski, editors, \emph{Machine Learning: ECML 2003}, pages 361--372, Berlin, Heidelberg, 2003. Springer Berlin Heidelberg.
\newblock \doi{10.1007/978-3-540-39857-8_33}.

\bibitem[Singh et~al.(2019)Singh, Gehr, P\"{u}schel, and Vechev]{singh2019eth}
Gagandeep Singh, Timon Gehr, Markus P\"{u}schel, and Martin Vechev.
\newblock \emph{\protect{An abstract domain for certifying neural networks}}, volume~3.
\newblock Association for Computing Machinery, New York, NY, USA, jan 2019.
\newblock \doi{10.1145/3290354}.

\bibitem[Tao et~al.(2019)Tao, Gao, and Wang]{tao2019testing}
Chuanqi Tao, Jerry Gao, and Tiexin Wang.
\newblock \protect{Testing and Quality Validation for AI Software-Perspectives, Issues, and Practices}.
\newblock \emph{IEEE Access}, 7:\penalty0 120164--120175, 08 2019.
\newblock \doi{10.1109/ACCESS.2019.2937107}.

\bibitem[Veropoulos et~al.(1999)Veropoulos, Campbell, and Cristianini]{SVM_Veropoulos}
K.~Veropoulos, C.~Campbell, and N.~Cristianini.
\newblock \protect{Controlling the Sensitivity of Support Vector Machines}.
\newblock In \emph{Proceedings of the International Joint Conference on {AI}}, pages 55--60, 1999.

\bibitem[Wang et~al.(2021)Wang, Han, Li, Zhang, and Cheng]{9408661}
Le~Wang, Meng Han, Xiaojuan Li, Ni~Zhang, and Haodong Cheng.
\newblock \protect{Review of Classification Methods on Unbalanced Data Sets}.
\newblock \emph{IEEE Access}, 9:\penalty0 64606--64628, 2021.
\newblock \doi{10.1109/ACCESS.2021.3074243}.

\bibitem[Zhang et~al.(2021)Zhang, Xie, Bai, Yu, Li, and Gao]{zhang2021survey}
Chen Zhang, Yu~Xie, Hang Bai, Bin Yu, Weihong Li, and Yuan Gao.
\newblock \protect{A survey on federated learning}.
\newblock \emph{Knowledge-Based Systems}, 216:\penalty0 106775, 2021.
\newblock ISSN 0950-7051.
\newblock \doi{10.1016/j.knosys.2021.106775}.

\end{thebibliography}

\end{document}